\documentclass[fleqn,10pt]{wlscirep}
\usepackage[utf8]{inputenc}
\usepackage[T1]{fontenc}
\usepackage{soul}
\usepackage{graphicx}
\title{CoolMomentum: A Method for Stochastic Optimization by Langevin Dynamics with Simulated Annealing\footnote{This is a post-peer-review, precopyedit version of an article published in \textit{Scientific Reports}. The final authenticated version is available online at: \url{https://doi.org/10.1038/s41598-021-90144-3}}}

\author[1,*,+]{Oleksandr Borysenko}
\author[2,+]{Maksym Byshkin}

\affil[1]{National Science Center "Kharkiv Institute of Physics and Technology" Kharkiv, 61108, Ukraine}
\affil[2]{Institute of Computational Science, Università della Svizzera italiana, Lugano, 6900, Switzerland}

\affil[*]{alessandro.borisenko@gmail.com}

\affil[+]{these authors contributed equally to this work}


\begin{abstract}
Deep learning applications require global optimization of non-convex objective functions, which have multiple local minima. The same problem is often found in physical simulations 
and may be resolved by the methods of Langevin dynamics with Simulated Annealing, which 
is a well-established approach for minimization of many-particle potentials. 
This analogy provides useful insights for non-convex stochastic optimization in machine learning. Here we find that integration of the discretized Langevin equation gives a coordinate updating rule equivalent to the famous Momentum optimization algorithm. As a main result, we show that a gradual decrease of the momentum coefficient from the initial value close to unity until zero is equivalent to application of Simulated Annealing or slow cooling, in physical terms. Making use of this novel approach, we propose CoolMomentum --
 a new stochastic optimization method. Applying Coolmomentum to optimization of Resnet-20 on Cifar-10 dataset and Efficientnet-B0 on Imagenet, we demonstrate that it is able to achieve high accuracies. 
\end{abstract}
\begin{document}

\flushbottom
\maketitle
%
%
\thispagestyle{empty}


\section*{\label{Introduction}Introduction}

A rapid growth of machine learning applications has been observed in recent years.   
Training of machine learning models is performed by finding such values of their parameters that optimize an objective function. Usually the number of parameters is large and the training dataset is massive. The first order stochastic optimization methods are proved to be most appropriate in this case. To reduce computational costs, the gradient of the objective function with respect to the model parameters is computed on relatively small subsets of the training data, called mini-batches. The resulting value is an unbiased stochastic estimator of the true gradient and it is used with stochastic gradient descent (SGD) methods. 

Most theoretical works are focused on convex optimization \cite{schmidt2017minimizing,reddi2019convergence}, but optimization of nonconvex objective functions is required usually. Empirically it is shown that several optimization algorithms, e.g SGD with momentum \cite{Rumelhart1986}, Adagrad\cite{duchi2011adaptive}, RMSProp\cite{tieleman2012lecture}, Adadelta\cite{zeiler2012adadelta} and Adam\cite{kingma2014adam}  are efficient for training artificial neural networks and optimization of nonconvex objective functions\cite{goodfellow2016deep,bottou2018optimization}. In nonconvex setting, the objective function has multiple local minima and the efficient algorithms rely on the “hill climbing” heuristics. Currently, there is a significant gap between mathematical theory and heuristic stochastic optimization methods popular in machine learning.

There is a useful connection between multivariate optimization and molecular simulations. In molecular simulations the hill climbing heuristics is related to passing through the energy barriers. Local energy minima are typical for molecular systems. Based on the detailed analogy between the multivariate optimization and annealing in molecular systems, the Simulated Annealing method was proposed \cite{kirkpatrick1983optimization}. 
This nature-inspired optimization method takes name and inspiration coming from annealing (slow cooling) in materials science and computational physics. Simulation of annealing can be used to find an approximation of the global minimum for a function $U(x)$ of many variables. In physics this function is known as a potential energy $U(x)$ of a molecular system. In order to apply Simulated Annealing, one needs a method for sampling from the Gibbs-Boltzmann distribution

\begin{equation} \label{eq:Gibbs}
w_n=\exp(-U_n/T)/Z,
\end{equation}
where $T$ is a parameter called temperature and $Z$ is a normalizing constant, $Z=\sum_{n} \exp(-U_{n}/T)$. The Gibbs distribution $w_{n}$ gives the probability to find a system $x$ in a state $n$ with energy $U_{n}=U(x)$. The mean of any quantity $f(x)$ may be calculated utilising the Gibbs distribution, using the formula $\left \langle f  \right \rangle=\sum_{n}  w_{n} f$.  The Gibbs distribution is one of most important formulas in statistical physics \cite{landau1980course}.

Classical methods for simulation of molecular systems are Markov chain Monte Carlo (MCMC), molecular dynamics (MD) and Langevin dynamics (LD). Either MD, LD or MCMC lead to equilibrium averaged distributions in the limit of infinite time or number of steps. If simulation is performed at a constant temperature $T$, these methods may be used to generate samples of Eq.~(\ref{eq:Gibbs}). Simulated Annealing can be used with any of these methods, but instead of performing simulation at a constant temperature $T$, the temperature should be decreased slowly. By performing simulation first at high temperature and then gradually decreasing the temperature value, the states close to the global minimum of $U(x)$ may be found. MCMC, MD and LD have different application areas. MD and LD are based on a numerical integration of the classical equation of motion. They simulate the dynamics of systems, based on the values of the gradient $dU(x)/dx$, that has to be computed on every step. MCMC does not require the gradient information, only $U(x)$ values are required to compute the Metropolis acceptance probability. MCMC methods may overcome energy barriers more efficiently, but they require special MCMC proposals, and there are no equivalently efficient proposals for different systems. If the values of $dU(x)/dx$ are available, then MD and LD are more straightforward methods. 

The adaptation of MCMC and LD for optimization is a prospective research direction \cite{ma2019sampling}. MCMC methods are widely used in machine learning, but applications of Langevin dynamics to machine learning only start to appear \cite{welling2011bayesian,ding2014bayesian,NIPS2017_6664,ma2019there,wenzel2020good}. In this paper, we propose to adapt the methods of Langevin dynamics to the problems of nonconvex optimization, that appear in machine learning. In Section~\ref{Langevin} we give a brief review of the methods of Molecular and Langevin dynamics and show their relation to the stochastic optimization method. In Section~\ref{Optimization} we discuss the basics of Simulated Annealing. In Section~\ref{RelationMOM} we explore the relation of the discretized Langevin equation with the Momentum optimizer. In Section~\ref{OptimizationAlg} we present the details of the CoolMomentum algorithm. In Section~\ref{Experiments} we evaluate the new algorithm and compare its performance to Adam and Momentum and we leave Section ~\ref{Conclusions} for conclusions. 

\section{\label{Langevin}Molecular and Langevin Dynamics}

Molecular and Langevin dynamics were proposed for simulation of molecular systems by integration of the classical equation of motion to generate a trajectory of the system of particles. Both methods operate with the classical equation of motion of $N$ particles with coordinates $x=(x_1, x_2, .., x_N)$, velocities $v=dx/dt$ and accelerations $a=d^{2} x/dt^{2}$. The Newton’s equation of motion for a conservative system is given by

\begin{equation} \label{eq:Newton}
m\frac{d^{2} x}{dt^2}=f(x)\equiv  -\frac{dU(x)}{dx},
\end{equation}
where $m$ is the  mass of particles, $f(x)$ is known as force, and $U(x)$ is the potential energy. The kinetic energy is given by

\begin{equation} \label{eq:Ekin}
E_k=\sum_{i=1}^{N} \frac{m_{i} v_{i}^2}{2}.
\end{equation}

There are several integration schemes based on discretization of the differential equation~(\ref{eq:Newton}), the Verlet and Velocity-Verlet algorithms being the most popular among them \cite{schlick2010molecular}.

In conservative systems, described by Eq.~(\ref{eq:Newton}), the sum of potential and kinetic energies conserves:  $ E_k+U=const $. The mean double kinetic energy per dimension per particle  

\begin{equation} \label{eq:Temperature}
T_k=\frac{1}{3\cdot N} \left \langle \sum_{i=1}^{N}m_{i} v_{i}^2 \right \rangle =\frac{2\langle E_{k} \rangle}{3N}
\end{equation}
is a parameter called temperature. Here and below $\langle f \rangle = \frac{1}{t} \int_{0}^{t} f(t')dt'$ means averaging over time or iterations. 
Often it is desirable to perform simulations at a given temperature, so that

\begin{equation} \label{eq:Thermo}
T_k \approx T,
\end{equation}
where $T$ is the desirable temperature, a parameter of the simulation. In physical simulations, an algorithm or a rule which controls the temperature is conventionally called a thermostat.

If molecules under consideration are allowed to exchange their kinetic energy with a medium (other molecules), then their total energy does not conserve any more. 
 In Langevin Dynamics, two forces are added to the conservative  force to account for the energy exchange with the medium - a friction force proportional to the velocity with a friction coefficient $\gamma \geq 0$ and a thermal white noise. These two forces play a role of the thermostat in LD.
Explicitly, the Langevin dynamics may be described by the following  equation \cite{bussi2007accurate,vanden2006second, van1982algorithms,schlick2010molecular}:

\begin{equation} \label{eq:Langevin}
m\frac{d^{2} x}{dt^2}= f(x)-m\gamma v(t)+R(t), 
\end{equation}
where $R(t)$ is a random uncorrelated force with zero mean and a temperature-dependent magnitude: 

\begin{align} \label{eq:Noise}
\left \langle R(t) \right \rangle=0; \nonumber \\
\left \langle R(t)R(t') \right \rangle=2mT\gamma \delta (t-t'), 
\end{align} 
$\delta (t-t')$ being the Dirac Delta function.

The magnitude of the friction $\gamma $ determines the relative strength of the  dissipation forces with respect to the conservative force  $f(x)$. If $\gamma=0 $, one only has conservative forces without energy dissipation and Eq.~(\ref{eq:Langevin}) reduces to Eq.~(\ref{eq:Newton}).

Several discretization schemes for the Langevin equation were proposed, e.g. a generalization of the Velocity-Verlet integrator to Langevin Dynamics by Vanden-Eijnden and Cicotti \cite{vanden2006second}.

In the high friction limit, the acceleration term in the LHS of Eq.~(\ref{eq:Langevin}) may be neglected and one has

\begin{equation} \label{eq:Brownian}
m\gamma v(t)dt=f\left(x(t)\right)dt+R(t)dt. 
\end{equation}
It is known as overdamped Langevin equation. Its first order integrator was proposed by Ermak and McCammon  \cite{schlick2010molecular}:

\begin{equation} \label{eq:Over}
x(t+\Delta t)=x(t)+\Delta t\frac{1}{m\gamma }f\left(x(t)\right)+\sqrt{\Delta t}\sqrt{\frac{2T}{m\gamma}}\xi ,
\end{equation}
where $\xi$ is a random Gaussian noise with zero mean and unit variance. The last term in the RHS of Eq.~(\ref{eq:Over}) 
results from the integral of the random force~(\ref{eq:Noise}) $\int_{0}^{\Delta t} R(t')dt'$, known as the Wiener process.

From Eq.~(\ref{eq:Over}) one can see that $\gamma$  enters its denominator, and would result in infinitely large values of updating steps if friction $\gamma$ is close to zero. Therefore, this integrator is appropriate for essentially high friction values only.

\section{\label{Optimization}Optimization by Simulated Annealing for Machine Learning}

Simulated Annealing (SA) is a well established optimization technique to locate the global $U(x)$ minimum without getting trapped into local minima. Though originally SA was proposed as an extension of MCMC \cite{kirkpatrick1983optimization}, SA can be considered as an extension of either MCMC or molecular/Langevin dynamics (see Ch. 12.5 of Schlick\cite{schlick2010molecular}). In this paper we propose to adapt these methods to the problem of optimization in machine learning, that require minimization of a function based on the values of its gradients. For instance, this function may be attributed as a loss and the values of the gradient $dU/dx$ may be computed by backpropagation~\cite{Rumelhart1986}.

To get an idea about the basics of Simulated Annealing, one can think as follows. Consider a heavy ball moving in a one-dimensional potential well with multiple minima, separated by barriers. The deepest of the minima is the global one, the others are local. Let the initial mean kinetic energy of the ball be high enough to overcome any energy barrier, therefore the ball passes through all the minima on its quasiperiodic trajectory. According to Eq.~(\ref{eq:Temperature}), high kinetic energy corresponds to high temperature. Suppose now, that the temperature (mean kinetic energy) is gradually decreased. This process has to be slow enough, to ensure that the characteristic cooling time is much longer than the characteristic time of the quasiperiodic motion. In the course of this cooling, another higher-lying local minimum eventually becomes inaccessible as soon as the mean kinetic energy becomes less than the height of its energy barrier. And finally, when the mean kinetic energy becomes less than the barrier between the global and the first local minimum, the ball becomes localized in the global minimum. This consideration may be freely generalized to multiple dimensions.

 Therefore, if the values of $ dU(x)/dx$ are available, then Simulated Annealing in a combination with molecular dynamics is a well-established method for locating the global minimum of a multivariate function $U(x)$.  It is proved to be particularly efficient for nonconvex functions. The value of constant $m$ may be selected arbitrary. For simplicity we can set $m=1$ throughout. SA may be implemented using  e.g. the Velocity-Verlet integrator and one of the thermostats \cite{schlick2010molecular}. The beauty of the described above  SA  is that it has theoretical  guarantees to converge to the global minimum of a nonconvex function \cite{granville1994simulated}. However, the convergence is guaranteed in the limit of very slow cooling only. In practice, the efficiency of SA depends on the annealing schedule, that has to be specified by the user.

If the training data is large, then it is computationally expensive to compute the loss and its gradient on the full training set. In this case stochastic optimization is proved to be the only appropriate approach. In stochastic optimization, the values of the loss and its gradient are estimated approximately, on small subsets of training data, called minibatches. If these minibatches are selected randomly from the training data, then the estimated values of the loss $\hat{U}(x)$ and its gradient $d\hat{U}/dx$ are the Monte Carlo approximations of their exact values. Stochastic Gradient Descent is the simplest optimization method and is the method of choice for many applications. Formally it may be written as

\begin{equation} \label{eq:SGD}
x_{n+1}=x_{n}- lr \frac{d\hat{U} }{dx}.
\end{equation}
In Eq.~(\ref{eq:SGD}) the constant $lr$ is known as a learning rate, and $d\hat{U}/dx$ is a stochastic gradient. This equation can be compared with Eq.~(\ref{eq:Over}). Besides the thermal noise, there are only two differences between these equations: I) $f(x)=-\frac{dU}{dx}$  in (\ref{eq:Over}) is the exact gradient, while $\frac{d\hat{U} }{dx}$ in (\ref{eq:SGD}) is the stochastic gradient and II) the discrete time variable $t$  in Eq.~(\ref{eq:Over}) is substituted with the iteration number $n$, so that $lr=\Delta t/(m\gamma )$.  

Though the Monte Carlo approximation $d\hat{U}/dx$ is a good unbiased approximation, it is still an approximation and contains noise. One can write  \cite{friedlander2012hybrid}

\begin{equation} \label{eq:RandF}
\hat{f}=-d\hat{U}/dx=-d{U}/dx+R,
\end{equation}
where $R$ is an uncorrelated random noise with zero mean.  
If the size of the minibatch is large, or the gradient $d\hat{U}/dx$ is computed on the full training data set, then $d\hat{U}/dx=d{U}/dx$ and $R=0$. 
In this case molecular dynamics in a combination with simulated annealing is a well established method for global optimization \cite{schlick2010molecular}. 
On the other hand, if the batch size is small, then the random noise $R$ may be large. In this case the Langevin dynamics 
in a combination with simulated annealing 
may be adapted for global optimization \cite{schlick2010molecular}. 

\section{\label{RelationMOM}Relation of the Langevin equation with Momentum optimizer}
Setting $m=1$ in the Langevin equation~(\ref{eq:Langevin}) and defining the stochastic force $\hat{f}=f+R$, one obtains
\begin{equation} \label{eq:Langevin_fin1}
\frac{\Delta^{2}x}{\Delta t^{2}}= \hat{f} - \gamma v(t).
\end{equation}
Expressing the time derivatives in finite differences, one can obtain the next equation:

\begin{equation} \label{eq:Langevin_fin2}
\frac{\Delta^{2}x}{\Delta t^{2}}=\frac{\Delta x_{n+1} - \Delta x_{n}}{\Delta t^{2}} = \hat{f}_{n} - \gamma \frac{\Delta x_{n+1} + \Delta x_{n}}{2 \Delta t}.
\end{equation}

Now, it is straightforward to obtain the next coordinate updating formula:

\begin{equation} \label{eq:momentum}
\Delta x_{n+1} = \rho \Delta x_{n} + \hat{f}_{n} \cdot lr                     
\end{equation}
with

\begin{equation} \label{eq:rho}
\rho = \frac{1-\gamma \Delta t /2}{1+\gamma \Delta t /2}
\end{equation}
and

\begin{equation} \label{eq:lr}
lr = \frac{\Delta t^{2}}{1+\gamma \Delta t /2} = \frac{1+\rho}{2}\Delta t^{2}.
\end{equation}

Eq.~(\ref{eq:momentum}) is nothing else but a famous Momentum optimization algorithm \cite{Rumelhart1986} with $\rho$ being a momentum coefficient and $lr$ a learning rate constant. 

Due to the change to discrete variables and $m=1$, Eq.~(\ref{eq:Noise}) becomes:

\begin{align} \label{eq:Thermostat_gamma}
\left \langle R_{n} \right \rangle=0; \nonumber \\
\left \langle R^{2}_{n} \right \rangle \Delta t = 2\gamma T.
\end{align}
Using Eq.~(\ref{eq:rho}) to obtain 

\begin{equation} \label{eq:gamma}
\gamma = \frac{2}{\Delta t} \cdot \frac{1-\rho}{1+\rho},
\end{equation}
one can change the last Eq.~(\ref{eq:Thermostat_gamma}) to:

\begin{equation} \label{eq:Thermostat_rho}
\left \langle R^{2}_{n} \right \rangle \Delta t^{2} = 4T\cdot \frac{1-\rho}{1+\rho}.
\end{equation}
For many machine learning applications the optimal $\rho$ value is in the range from 0.5 to 0.99. If  $\rho=0$ then Eq.~(\ref{eq:momentum}) becomes equivalent to  Eq.~(\ref{eq:SGD}), the Langevin dynamics becomes overdamped, and the Momentum optimizer becomes SGD. 

\section{\label{OptimizationAlg}Algorithm}

In order to apply Simulated Annealing for optimization, one needs a thermostat to control the temperature. In addition, a temperature schedule (or cooling strategy) has to be specified by the user. The temperature itself does not enter explicitly into our algorithm described by Eqs.~(\ref{eq:momentum})-(\ref{eq:lr}) (see also pseudocode in Table~\ref{Alg}). From Eq.~(\ref{eq:Thermostat_rho}) one can see that, for $\left \langle R^{2}_{n} \right \rangle \Delta t^{2} = const$, the product of the temperature and a function of the momentum coefficient stays constant: $4T(1-\rho)/(1+\rho) = const $. Therefore, instead of decreasing the temperature directly, one can increase the ratio $(1-\rho)/(1+\rho)$ by decreasing the momentum coefficient $\rho$, which enters our algorithm explicitly.

From Eqs.~(\ref{eq:rho}) and~(\ref{eq:gamma})  one can see that $\rho$ decreases from unity to zero as $\gamma$ increases from zero to its maximal value $2/\Delta t$, which corresponds to the overdamped regime.  
The decreasing $\rho$ schedule has to be specified by the user. Different $\rho$ schedules may be used. A possible $\rho$ schedule is given by

\begin{equation} \label{eq:cooling}
\rho_{n}=1-(1-\rho_{0})/\alpha^{n}.
\end{equation}
If $\alpha=1$ then $\rho_{n}=\rho_{0}$, and if $\alpha<1$ then $\rho_{n}$ is a decreasing function of $n$. In the Momentum optimizer the $\rho_{n}$ value should be in the range from 0 to 1. Let $S$ be the number of steps (usually $S$= number of epochs $\cdot$ steps per epoch). Then the algorithm we propose may be presented as a pseudocode given in Table~\ref{Alg}.

\begin{table}[hbt!] 
  \caption{\label{Alg}   } 
  \centering
  \begin{tabular}{lll}
    \toprule
    \multicolumn{2}{l}{\textbf{Algorithm "CoolMomentum"}}         \\
 \midrule
& Require: $lr=\Delta t^{2}$ (base learning rate) \\
& Require: $\rho_{0}$ (initial momentum coefficient) \\
& Require: $S$ (number of iterations) \\
& Compute: $\alpha=(1-\rho_0)^{1/S}$ (cooling rate) \\
& Initialization: $x_0$ (Initial parameter vector) \\
& Initialization: $\Delta x_{0} = 0$ (Initialize update vector)  \\

& \textbf{for} $n = 0 .. (S-1)$  \textbf{do}:  (loop over S iterations) \\
& \hspace{5mm} $\hat{f}(x_{n})=-d\hat{U}/dx$ (compute stochastic gradient) \\
& \hspace{5mm} $\rho_{n}=\max \left( 0,1-(1-\rho_{0})/\alpha^{n} \right)$ (slowly decrease $\rho$ value until zero)\\
& \hspace{5mm} $lr_{n}=lr\cdot \left(1 + \rho_{n} \right)/2$ (recalculate the learning rate)\\
& \hspace{5mm}$\Delta x_{n+1}=\rho_{n}  \Delta x_{n} + \hat{f}(x_{n})\cdot lr_{n}$ (update momentum)\\ 
& \hspace{5mm}$x_{n+1}=x_{n} + \Delta x_{n+1}$ (update parameters) \\
& \textbf{end do} \\
& return $x_{S}$ (Resulting parameters) \\
    \bottomrule
  \end{tabular}
\end{table}

Comparing with the classical Momentum optimizer, described by Eq.~(\ref{eq:momentum}), this algorithm requires one additional hyperparameter $\alpha$, that we call a "cooling rate".
Every additional hyperparameter may be painful for machine learning application. However, a good $\alpha$ value may be easily computed. In  Simulated Annealing the temperature should be slowly decreased until some minimal value, and therefore the $\rho$ value should be slowly decreased until $\rho=0$. Given $\rho_S=0$, from Eq.~(\ref{eq:cooling}) one can obtain:

\begin{equation} \label{eq:cooling_rate}
\alpha=(1-\rho_0)^{1/S}.
\end{equation}

\section{\label{Experiments}Evaluation}
To evaluate our optimization method, we study the problem of image classification. We trained a deep residual neural network \cite{he2016deep} ResNet-20 on the CIFAR-10 dataset with 50000 training images and 10000 testing ones using Adam~\cite{kingma2014adam}, Momentum~\cite{Rumelhart1986} and Coolmomentum optimizers. This model has a complicated architecture, more than 270k of trainable parameters and therefore it is a good model to check the performance of optimization methods. We used the code shared by the Keras team~\cite{Resnet-20}. Training of this model for 200 epochs on gtx1080ti GPU takes about 2 hours. For the Adam optimizer we took the initial value of the learning rate $lr=0.001$ with an original learning rate decay schedule, $\beta_1=0.9$ and $\beta_2=0.999$. For the Momentum optimizer we took the initial value of the learning rate $lr=0.01$ with an original learning rate decay schedule and $\rho=0.9$ for the momentum coefficient. For Adam and Momentum the learning rate decay factor of $0.1$ was applied after the 80th, 120th and 160th epochs and a factor of $0.5$ was applied after the 180th epoch. For Coolmomentum we took the base value of the learning rate $lr=0.01$ and the value of the cooling rate $\alpha$ was taken from Eq.~(\ref{eq:cooling_rate}) with $\rho_{0}=0.99$. The values of hyperparameters were selected by the trial and error method (see Table~\ref{tab:hyper1}). For the sake of reproducability, all calculations were performed with the same fixed random generator's seed value.

\begin{figure}[ht]
\centering
\includegraphics[width=\linewidth]{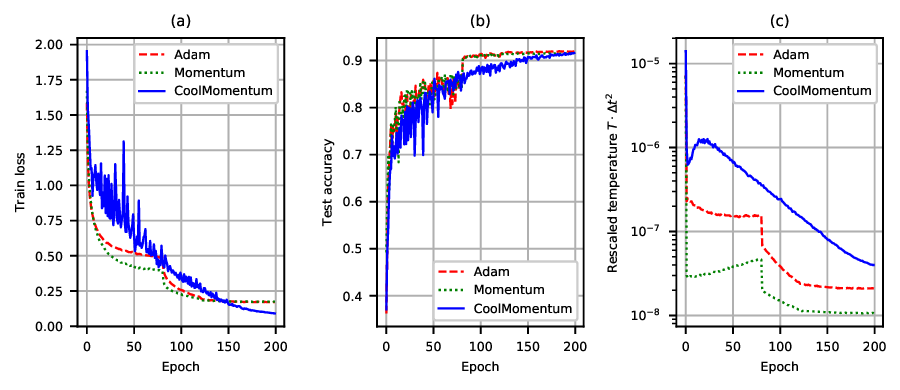}
\caption{Cifar-10 classification with ResNet-20: training loss (a), test accuracy (b) and rescaled temperature ($T\cdot \Delta t^2$) (c)}
\label{Fig:1}
\end{figure}

In order to check the performance of the optimization methods on ResNet-20, for each epoch we compute the training loss on the training data set (50000 images) and the testing accuracy on the testing data set (10000 images), and compare the optimization results in Fig.~\ref{Fig:1}~(a) and~(b), respectively. 

To be sure that Simulated Annealing is applied properly, i.e. that the temperature is decreased slowly, one needs a method to calculate the temperature directly during the optimization process. This may be done by using Eq.~(\ref{eq:Temperature}), setting $m=1$ and changing to discrete variables to obtain:

\begin{equation} \label{eq:Temperature2}
T=\frac{1}{\rm{Size}} \left \langle \sum_{i=1}^{\rm{Size}} v_{i}^2 \right \rangle = \frac{1}{\rm{Size} \cdot S} \sum_{i=1}^{\rm{Size}} \sum_{n=1}^{S} \left(\frac{\Delta x_{i, n}}{\Delta t}\right)^2, 
\end{equation}
where $\rm{Size}$ is a number of training parameters of the model and $S$ is a number of time iterations per epoch.

In Fig.~\ref{Fig:1}~(c) we present the values of rescaled temperature $T \cdot \Delta t^{2}$ calculated with Eq.~(\ref{eq:Temperature2}) for all the three optimizers being compared. We choose to calculate rescaled temperature instead of the ordinary one because the actual value of the time step $\Delta t$ is inavailable for Adam. From Fig.~\ref{Fig:1}~(c) one can see that on the first epoch the temperature significantly drops down for all three optimizers, but only in the case of Coolmomentum it evolves continuously on further epochs, while it changes stepwise according to the prescribed learning rate decay schedule for Adam and Momentum. Therefore, Coolmomentum performs optimization in the Simulated Annealing regime, and by slowly decreasing the temperature it samples the states of the Gibbs distribution~(\ref{eq:Gibbs}), which continuously approach the global minimum of the loss function. On the contrary, Adam and Momentum drop the temperature in a stepwise manner. In materials science and physical simulations this cooling regime is called quenching. It produces a variety of non-equilibrium disordered structures, including different glasses. Similarly to physical systems, in this regime the trained model becomes caught in a local minimum of the loss function, and continues to walk there, because the temperature is too low to overcome the local barrier. Indeed, from Fig.~\ref{Fig:1}~(a) one can see that both Adam and Momentum saturate to the constant (and equal) value of the training loss, while Coolmomentum continuously goes below this level.   

On the first epochs the training and testing results, produced by CoolMomentum, are worse than those of Momentum and Adam. Indeed, on the first epochs Coolmomentum gives the temperature values significantly higher than Adam and Momentum do (see Fig.~\ref{Fig:1}~(c)). But at high temperatures the Gibbs distribution (\ref{eq:Gibbs}) is less efficient to distinguish between the states with high and low values of the loss function. Nevertheless, as the temperature decreases, Coolmomentum achieves the top values produced by others in terms of the test accuracy (see Fig.~\ref{Fig:1}~(b)) and outperforms them in terms of training loss values (see Fig.~\ref{Fig:1}~(a)), which encourages further studies of different models and datasets.

\begin{table}[ht]
\centering
\begin{tabular}{|l|l|l|l|}
\hline
$\rho_0 \backslash lr$   & 0.001 & 0.01 & 0.02 \\
\hline
0.9 & 0.8697 & 0.9062 & 0.9139 \\
\hline
0.99 & 0.8972 & \bf{0.9160} & 0.9057 \\
\hline
0.999 & 0.9064 & div & div \\
\hline
\end{tabular}
\caption{\label{tab:hyper1} Test accuracy of Resnet-20, trained on Cifar-10 for 200 epochs vs. Coolmomentum hyperparameters $lr$ and $\rho_0$. "div" means "divergent". The best value is in bold.}
\end{table}

\begin{figure}[ht]
\centering
\includegraphics[width=\linewidth]{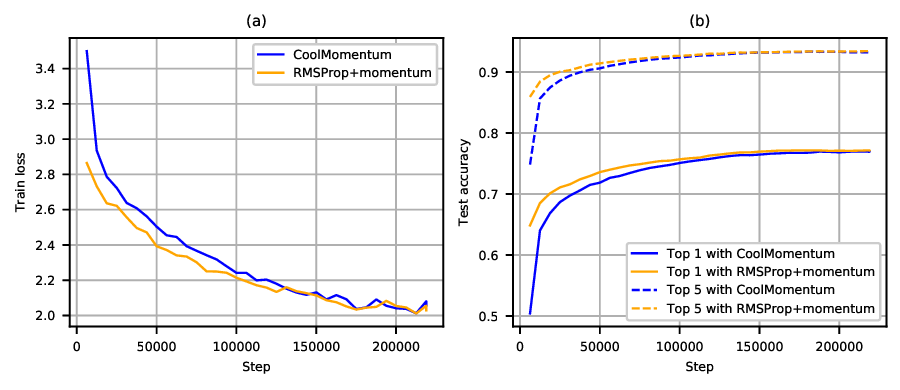}
\caption{Imagenet classification with Efficientnet-B0: training loss (a) and test accuracy (b)}
\label{Fig:2}
\end{figure}

We also trained Efficientnet~\cite{tan2020efficientnet} B0 on the Imagenet (1000 classes) dataset~\cite{Imagenet} with $1281167$ training images and $50000$ testing ones for $218948$ steps (about $350$ epochs) with batch size $2048$ for about 30 hours on v2-8 cloud TPU. At first we ran the publicly available code for training Efficientnet on cloud TPU~\cite{Efficientnet-tutorial} with default settings: RMSprop with batch-scaled learning rate $0.128=0.016\cdot(2048/256)$, momentum coefficient $0.9$, exponential running average decay $0.9$, $\epsilon=0.001$, learning rate decay factor $0.97$ for each $2.4$ epochs with a linear warm-up for the first 5 epochs. Then we modified it to realize Coolmomentum with base $lr = 0.6$, $\rho_{0}=0.99$ and the cooling rate $\alpha$ calculated from Eq.~(\ref{eq:cooling_rate}). We set $\rho=0$ for the first 5 epoch for warm-up. The value of the base learning rate was selected by the trial and error method based on the data of Table~\ref{tab:hyper2}.   

\begin{table}[ht]
\centering
\begin{tabular}{|l|c|c|c|c|}
\hline
$lr$           & 0.1&    0.2&    0.6&         0.7  \\
\hline
Top-1, \%    & 76.03&  76.55&  \bf{76.99}&  76.87\\
\hline
Top-5, \%    & 92.63&  93.08&  93.24&       \bf{93.32}  \\
\hline
\end{tabular}

\caption{\label{tab:hyper2} Top accuracies for Imagenet classification with Efficientnet-B0 optimized with Coolmomentum at different learning rates. The best values are in bold.}
\end{table} 

The results are presented in Fig.~\ref{Fig:2}. One can see that in this case Coolmomentum also achieves the top results.

\section{\label{Conclusions}Conclusions}

We explore relations between the Langevin dynamics and the stochastic optimization methods, popular in machine learning. The relation of underdamped Langevin dynamics with the Momentum optimizer was studied recently \cite{ma2019there}. In this paper we combine Langevin dynamics with Simulated Annealing. To apply Simulated Annealing, the temperature should be decreased slowly until some minimal value. This is usually done by decreasing the learning rate with a certain schedule. Indeed, from Eq.~(\ref{eq:Thermostat_rho}) one can see that, from decreasing the value of $lr\sim \Delta t^2$, the temperature $T$ decreases proportionally. Alternatively, we propose to adapt Simulated Annealing 
by slowly decreasing the momentum coefficient of the Momentum optimizer, and propose a decreasing schedule for the values of this coefficient.
In our case, 
at the minimal temperature the momentum coefficient becomes zero 
and the Langevin dynamics becomes overdamped. 

The proposed Coolmomentum optimizer requires only 3 tunable hyperparameters (base learning rate, initial momentum coefficient and the total number of optimization steps), while SGD with momentum requires 1 more parameter (learning rate decay factor) and RMSprop and Adam require 2 extra parameters (exponential running average coefficient and a small constant to avoid divergence). In this way, our approach is advantageous, because it reduces the number of tunable hyperparameters and, therefore, demands less computational budget to choose the best values~\cite{sivaprasad2020optimizer}.
We demonstrate that training of Resnet-20 on Cifar-10 dataset and Efficientnet-B0 on Imagenet with Coolmomentum optimizer allows to achieve high accuracies.
The obtained results indicate that the combination of the Langevin dynamics with Simulated Annealing is an efficient approach for gradient-based optimization of stochastic objective functions.

The convergence analysis of Simulated Annealing was performed by several authors \cite{geman1986diffusions,gidas1985global,gidas1985nonstationary,holley1988simulated,marquez1997convergence}. We hope to attract attention of researchers to this optimization method.

\bibliography{Coolmomentum}

\begin{thebibliography}{10}
\urlstyle{rm}
\expandafter\ifx\csname url\endcsname\relax
  \def\url#1{\texttt{#1}}\fi
\expandafter\ifx\csname urlprefix\endcsname\relax\def\urlprefix{URL }\fi
\expandafter\ifx\csname doiprefix\endcsname\relax\def\doiprefix{DOI: }\fi
\providecommand{\bibinfo}[2]{#2}
\providecommand{\eprint}[2][]{\url{#2}}

\bibitem{schmidt2017minimizing}
\bibinfo{author}{Schmidt, M.}, \bibinfo{author}{Le~Roux, N.} \&
  \bibinfo{author}{Bach, F.}
\newblock \bibinfo{journal}{\bibinfo{title}{Minimizing finite sums with the
  stochastic average gradient}}.
\newblock {\emph{\JournalTitle{Mathematical Programming}}}
  \textbf{\bibinfo{volume}{162}}, \bibinfo{pages}{83--112}
  (\bibinfo{year}{2017}).

\bibitem{reddi2019convergence}
\bibinfo{author}{Reddi, S.~J.}, \bibinfo{author}{Kale, S.} \&
  \bibinfo{author}{Kumar, S.}
\newblock \bibinfo{journal}{\bibinfo{title}{On the convergence of {a}dam and
  beyond}}.
\newblock {\emph{\JournalTitle{arXiv preprint arXiv:1904.09237}}}
  (\bibinfo{year}{2019}).

\bibitem{Rumelhart1986}
\bibinfo{author}{Rumelhart, D.}, \bibinfo{author}{Hinton, G.} \&
  \bibinfo{author}{Williams, R.}
\newblock \bibinfo{journal}{\bibinfo{title}{Learning representations by
  back-propagating errors}}.
\newblock {\emph{\JournalTitle{Nature}}} \textbf{\bibinfo{volume}{323}},
  \bibinfo{pages}{533--536} (\bibinfo{year}{1986}).

\bibitem{duchi2011adaptive}
\bibinfo{author}{Duchi, J.}, \bibinfo{author}{Hazan, E.} \&
  \bibinfo{author}{Singer, Y.}
\newblock \bibinfo{journal}{\bibinfo{title}{Adaptive subgradient methods for
  online learning and stochastic optimization.}}
\newblock {\emph{\JournalTitle{Journal of Machine Learning Research}}}
  \textbf{\bibinfo{volume}{12}} (\bibinfo{year}{2011}).

\bibitem{tieleman2012lecture}
\bibinfo{author}{Tieleman, T.} \& \bibinfo{author}{Hinton, G.}
\newblock \bibinfo{journal}{\bibinfo{title}{Lecture 6.5-rmsprop: Divide the
  gradient by a running average of its recent magnitude}}.
\newblock {\emph{\JournalTitle{COURSERA: Neural networks for machine
  learning}}} \textbf{\bibinfo{volume}{4}}, \bibinfo{pages}{26--31}
  (\bibinfo{year}{2012}).

\bibitem{zeiler2012adadelta}
\bibinfo{author}{Zeiler, M.~D.}
\newblock \bibinfo{journal}{\bibinfo{title}{{A}dadelta: an adaptive learning
  rate method}}.
\newblock {\emph{\JournalTitle{arXiv preprint arXiv:1212.5701}}}
  (\bibinfo{year}{2012}).

\bibitem{kingma2014adam}
\bibinfo{author}{Kingma, D.~P.} \& \bibinfo{author}{Ba, J.}
\newblock \bibinfo{journal}{\bibinfo{title}{Adam: A method for stochastic
  optimization}}.
\newblock {\emph{\JournalTitle{arXiv preprint arXiv:1412.6980}}}
  (\bibinfo{year}{2014}).

\bibitem{goodfellow2016deep}
\bibinfo{author}{Goodfellow, I.}, \bibinfo{author}{Bengio, Y.} \&
  \bibinfo{author}{Courville, A.}
\newblock \emph{\bibinfo{title}{Deep learning}} (\bibinfo{publisher}{MIT
  press}, \bibinfo{year}{2016}).

\bibitem{bottou2018optimization}
\bibinfo{author}{Bottou, L.}, \bibinfo{author}{Curtis, F.~E.} \&
  \bibinfo{author}{Nocedal, J.}
\newblock \bibinfo{journal}{\bibinfo{title}{Optimization methods for
  large-scale machine learning}}.
\newblock {\emph{\JournalTitle{Siam Review}}} \textbf{\bibinfo{volume}{60}},
  \bibinfo{pages}{223--311} (\bibinfo{year}{2018}).

\bibitem{kirkpatrick1983optimization}
\bibinfo{author}{Kirkpatrick, S.}, \bibinfo{author}{Gelatt, C.~D.} \&
  \bibinfo{author}{Vecchi, M.~P.}
\newblock \bibinfo{journal}{\bibinfo{title}{Optimization by simulated
  annealing}}.
\newblock {\emph{\JournalTitle{Science}}} \textbf{\bibinfo{volume}{220}},
  \bibinfo{pages}{671--680} (\bibinfo{year}{1983}).

\bibitem{landau1980course}
\bibinfo{author}{Landau, L.~D.} \& \bibinfo{author}{Lifshitz, E.~M.}
\newblock \emph{\bibinfo{title}{Course of theoretical physics}}, vol.
  \bibinfo{volume}{5. Statistical physics} (\bibinfo{publisher}{Pegamon},
  \bibinfo{year}{1980}).

\bibitem{ma2019sampling}
\bibinfo{author}{Ma, Y.-A.}, \bibinfo{author}{Chen, Y.}, \bibinfo{author}{Jin,
  C.}, \bibinfo{author}{Flammarion, N.} \& \bibinfo{author}{Jordan, M.~I.}
\newblock \bibinfo{journal}{\bibinfo{title}{Sampling can be faster than
  optimization}}.
\newblock {\emph{\JournalTitle{Proceedings of the National Academy of
  Sciences}}} \textbf{\bibinfo{volume}{116}}, \bibinfo{pages}{20881--20885}
  (\bibinfo{year}{2019}).

\bibitem{welling2011bayesian}
\bibinfo{author}{Welling, M.} \& \bibinfo{author}{Teh, Y.~W.}
\newblock \bibinfo{title}{Bayesian learning via stochastic gradient {L}angevin
  dynamics}.
\newblock In \emph{\bibinfo{booktitle}{Proceedings of the 28th international
  conference on machine learning (ICML-11)}}, \bibinfo{pages}{681--688}
  (\bibinfo{year}{2011}).

\bibitem{ding2014bayesian}
\bibinfo{author}{Ding, N.} \emph{et~al.}
\newblock \bibinfo{title}{Bayesian sampling using stochastic gradient
  thermostats}.
\newblock In \emph{\bibinfo{booktitle}{Advances in Neural Information
  Processing Systems}} (\bibinfo{year}{2014}).

\bibitem{NIPS2017_6664}
\bibinfo{author}{Ye, N.}, \bibinfo{author}{Zhu, Z.} \&
  \bibinfo{author}{Mantiuk, R.}
\newblock \bibinfo{title}{Langevin dynamics with continuous tempering for
  training deep neural networks}.
\newblock In \emph{\bibinfo{booktitle}{Advances in Neural Information
  Processing Systems}}, \bibinfo{pages}{618--626} (\bibinfo{year}{2017}).

\bibitem{ma2019there}
\bibinfo{author}{Ma, Y.-A.} \emph{et~al.}
\newblock \bibinfo{journal}{\bibinfo{title}{Is there an analog of {N}esterov
  acceleration for {MCMC}?}}
\newblock {\emph{\JournalTitle{arXiv preprint arXiv:1902.00996}}}
  (\bibinfo{year}{2019}).

\bibitem{wenzel2020good}
\bibinfo{author}{Wenzel, F.} \emph{et~al.}
\newblock \bibinfo{journal}{\bibinfo{title}{How good is the bayes posterior in
  deep neural networks really?}}
\newblock {\emph{\JournalTitle{arXiv preprint arXiv:2002.02405}}}
  (\bibinfo{year}{2020}).

\bibitem{schlick2010molecular}
\bibinfo{author}{Schlick, T.}
\newblock \emph{\bibinfo{title}{Molecular modeling and simulation: an
  interdisciplinary guide}}, vol.~\bibinfo{volume}{21}
  (\bibinfo{publisher}{Springer Science \& Business Media},
  \bibinfo{year}{2010}).

\bibitem{bussi2007accurate}
\bibinfo{author}{Bussi, G.} \& \bibinfo{author}{Parrinello, M.}
\newblock \bibinfo{journal}{\bibinfo{title}{Accurate sampling using {L}angevin
  dynamics}}.
\newblock {\emph{\JournalTitle{Physical Review E}}}
  \textbf{\bibinfo{volume}{75}}, \bibinfo{pages}{056707}
  (\bibinfo{year}{2007}).

\bibitem{vanden2006second}
\bibinfo{author}{Vanden-Eijnden, E.} \& \bibinfo{author}{Ciccotti, G.}
\newblock \bibinfo{journal}{\bibinfo{title}{Second-order integrators for
  {L}angevin equations with holonomic constraints}}.
\newblock {\emph{\JournalTitle{Chemical Physics Letters}}}
  \textbf{\bibinfo{volume}{429}}, \bibinfo{pages}{310--316}
  (\bibinfo{year}{2006}).

\bibitem{van1982algorithms}
\bibinfo{author}{Van~Gunsteren, W.} \& \bibinfo{author}{{B}erendsen, H.}
\newblock \bibinfo{journal}{\bibinfo{title}{Algorithms for {B}rownian
  dynamics}}.
\newblock {\emph{\JournalTitle{Molecular Physics}}}
  \textbf{\bibinfo{volume}{45}}, \bibinfo{pages}{637--647}
  (\bibinfo{year}{1982}).

\bibitem{granville1994simulated}
\bibinfo{author}{Granville, V.}, \bibinfo{author}{Kriv{\'a}nek, M.} \&
  \bibinfo{author}{Rasson, J.-P.}
\newblock \bibinfo{journal}{\bibinfo{title}{Simulated annealing: A proof of
  convergence}}.
\newblock {\emph{\JournalTitle{IEEE transactions on pattern analysis and
  machine intelligence}}} \textbf{\bibinfo{volume}{16}},
  \bibinfo{pages}{652--656} (\bibinfo{year}{1994}).

\bibitem{friedlander2012hybrid}
\bibinfo{author}{Friedlander, M.~P.} \& \bibinfo{author}{Schmidt, M.}
\newblock \bibinfo{journal}{\bibinfo{title}{Hybrid deterministic-stochastic
  methods for data fitting}}.
\newblock {\emph{\JournalTitle{SIAM Journal on Scientific Computing}}}
  \textbf{\bibinfo{volume}{34}}, \bibinfo{pages}{A1380--A1405}
  (\bibinfo{year}{2012}).

\bibitem{he2016deep}
\bibinfo{author}{He, K.}, \bibinfo{author}{Zhang, X.}, \bibinfo{author}{Ren,
  S.} \& \bibinfo{author}{Sun, J.}
\newblock \bibinfo{title}{Deep residual learning for image recognition}.
\newblock In \emph{\bibinfo{booktitle}{Proceedings of the IEEE conference on
  computer vision and pattern recognition}}, \bibinfo{pages}{770--778}
  (\bibinfo{year}{2016}).

\bibitem{Resnet-20}
\bibinfo{howpublished}{\url{https://keras.io/zh/examples/cifar10_resnet/}}.

\bibitem{tan2020efficientnet}
\bibinfo{author}{Tan, M.} \& \bibinfo{author}{Le, Q.~V.}
\newblock \bibinfo{journal}{\bibinfo{title}{Efficientnet: Rethinking model
  scaling for convolutional neural networks}}.
\newblock {\emph{\JournalTitle{arXiv preprint arXiv:1905.11946}}}
  (\bibinfo{year}{2020}).

\bibitem{Imagenet}
\bibinfo{howpublished}{\url{http://www.image-net.org/challenges/LSVRC/2012/downloads}}.

\bibitem{Efficientnet-tutorial}
\bibinfo{howpublished}{\url{https://cloud.google.com/tpu/docs/tutorials/efficientnet}}.

\bibitem{sivaprasad2020optimizer}
\bibinfo{author}{Sivaprasad, P.~T.}, \bibinfo{author}{Mai, F.},
  \bibinfo{author}{Vogels, T.}, \bibinfo{author}{Jaggi, M.} \&
  \bibinfo{author}{Fleuret, F.}
\newblock \bibinfo{journal}{\bibinfo{title}{Optimizer benchmarking needs to
  account for hyperparameter tuning}}.
\newblock {\emph{\JournalTitle{arXiv preprint arXiv:1910.11758}}}
  (\bibinfo{year}{2020}).

\bibitem{geman1986diffusions}
\bibinfo{author}{Geman, S.} \& \bibinfo{author}{Hwang, C.-R.}
\newblock \bibinfo{journal}{\bibinfo{title}{Diffusions for global
  optimization}}.
\newblock {\emph{\JournalTitle{SIAM Journal on Control and Optimization}}}
  \textbf{\bibinfo{volume}{24}}, \bibinfo{pages}{1031--1043}
  (\bibinfo{year}{1986}).

\bibitem{gidas1985global}
\bibinfo{author}{Gidas, B.}
\newblock \bibinfo{title}{Global optimization via the {L}angevin equation}.
\newblock In \emph{\bibinfo{booktitle}{1985 24th IEEE Conference on Decision
  and Control}}, \bibinfo{pages}{774--778} (\bibinfo{organization}{IEEE},
  \bibinfo{year}{1985}).

\bibitem{gidas1985nonstationary}
\bibinfo{author}{Gidas, B.}
\newblock \bibinfo{journal}{\bibinfo{title}{Nonstationary {M}arkov chains and
  convergence of the annealing algorithm}}.
\newblock {\emph{\JournalTitle{Journal of Statistical Physics}}}
  \textbf{\bibinfo{volume}{39}}, \bibinfo{pages}{73--131}
  (\bibinfo{year}{1985}).

\bibitem{holley1988simulated}
\bibinfo{author}{Holley, R.} \& \bibinfo{author}{Stroock, D.}
\newblock \bibinfo{journal}{\bibinfo{title}{Simulated annealing via {S}obolev
  inequalities}}.
\newblock {\emph{\JournalTitle{Communications in Mathematical Physics}}}
  \textbf{\bibinfo{volume}{115}}, \bibinfo{pages}{553--569}
  (\bibinfo{year}{1988}).

\bibitem{marquez1997convergence}
\bibinfo{author}{M{\'a}rquez, D.}
\newblock \bibinfo{journal}{\bibinfo{title}{Convergence rates for annealing
  diffusion processes}}.
\newblock {\emph{\JournalTitle{The Annals of Applied Probability}}}
  \bibinfo{pages}{1118--1139} (\bibinfo{year}{1997}).

\end{thebibliography}

\section*{Acknowledgements}

MB thanks Swiss National Science Foundation, grant number 167326, National Research Program 75
(Big Data) for financial support. OB thanks National Academy of Science of Ukraine, grant number 0121U108687 for financial support. We thank Jeff Dean, TensorFlow Research Cloud and Google Cloud Research for high performance computational resources.

\section*{Author contributions statement}

O.B. and M.B. equally developed the theory, conducted the numerical calculations, analysed the results and prepared the manuscript. 

\section*{Additional information}

\textbf{Our open source code is available at} \url{https://github.com/borbysh/coolmomentum}; \textbf{Competing interests} The authors declare no competing interests. 


\end{document}